\documentclass{article}
\usepackage{ijcai18}
\usepackage{graphicx}
\usepackage[algoruled]{algorithm2e}

\usepackage{tabularx}
\usepackage{booktabs,multirow}
\usepackage{subfigure}
\usepackage{pgfplots}
\usepackage{tikz}	
\usepackage{tabu}
\usepackage{amsmath}

\usepackage{times}
\usepackage{xcolor}
\usepackage{soul}
\usepackage[utf8]{inputenc}
\usepackage[small]{caption}

\title{Multi-cell LSTM Based Neural Language Model
}

\author{
Thomas Cherian, 
Akshay Badola and 
Vineet Padmanabhan 
\\ 
School of Computer \& Information Sciences, University of Hyderabad, India\\
thoma@uohyd.ac.in,
badola@uohyd.ac.in,
vineetcs@uohyd.ernet.in
}

\begin{document}

\maketitle

\begin{abstract}
Language models, being at the heart of many NLP problems, are always of great interest to researchers. Neural language models come with the advantage of distributed representations and long range contexts. With its particular dynamics that allow the cycling of information within the network, `Recurrent neural network' (RNN) becomes an ideal paradigm for neural language modeling. Long Short-Term Memory (LSTM) architecture solves the inadequacies of the standard RNN in modeling long-range contexts.  In spite of a plethora of RNN variants, possibility to add multiple memory cells in LSTM nodes was seldom explored. Here we propose a multi-cell node architecture for LSTMs and study its applicability for neural language modeling. The proposed multi-cell LSTM language models outperform the state-of-the-art results on well-known Penn Treebank (PTB) setup.
\end{abstract}

\section{Introduction}

A language model is a function, or an algorithm to learn such a function that grasps the salient characteristics of the distribution of sequences of words in a natural language, allowing one to make probabilistic predictions for the next word given preceding ones. Language models are at the heart of many NLP tasks like Speech Recognition, Machine Translation, Handwriting Recognition, Parsing, Information Retrieval and Part of Speech tagging. N-gram based approaches are the most popular techniques for language modeling \cite{ngram_Manning:1999:FSN:311445,ngram_JelMer80}. They are based on the Markov's assumptions that the probability of occurrence of a particular word in the sequence depends only on the occurrence of previous $n-1$ words. Though successful, they are unable to make use of the longer contexts and the word similarities. N-gram based approaches can not look into the context beyond a smaller `n', say 3 in case of the trigram approach. Also they treat each word as a stand-alone entity and hence are unable to identify and utilize the similarity of words. The language models based on neural networks are termed as neural language models. These models exploit neural network's ability to learn distributed representations to fight the `curse of dimensionality' \cite{curseof_Bengio:2000:TCD:2325773.2326573,Bengio:2003:NPL:944919.944966}. Distributed representations also equip the neural models with the ability to utilize the word similarities. In contrast to the n-gram models; because of the use of recurrent neural networks; neural models do not require to predefine the context length. This allows the models to leverage larger contexts to make the predictions. These advantages make neural language models a popular choice despite the high computational complexity of the model.\\

``Recurrent neural networks form a very powerful and expressive family for sequential tasks. This is mainly because of the high dimensional hidden state with non-linear dynamics that enable them to remember and process past information." \cite{ICML2011Sutskever_524}. Cheap and easily computable gradients by means of back-propagation through time (BPTT) made them even more attractive. Even then RNNs failed to make their way into the mainstream research and applications for a long time due to the difficulty in effectively training them. The 'Vanishing and Exploding gradient' problems; discussed by Bengio et al. \cite{Bengio:1994:LLD:2325857.2328340} led to the complete negligence of the model for decades until recently. In 1997, Hochreiter et al. \cite{LSTM_Hochreiter:1997:LSM:1246443.1246450} came up with the LSTM architecture to deal with the inadequacies of the standard RNN in modeling long-range contexts. Gers et al. \cite{LSTM_GersF.:1999:LFC:870468} further enhanced the standard LSTM model with the addition of the forget gate. LSTMs prevent the `vanishing gradient problem' with the use of `Constant Error Carousels' (CECs) \cite{LSTM_Hochreiter:1997:LSM:1246443.1246450}; also known as `memory cells'. Currently LSTMs are at the core of RNN research. Many variants of basic LSTM have been proposed. However, to the best of our knowledge, these studies were centered around the efficient use of the gates in LSTM cell and none of them studied the effects of incorporating multiple memory cells into the single LSTM node. Here we move in this direction, propose a multi-cell LSTM architecture and studies its application on the neural language modeling. \\ 

We propose a multi-cell node architecture for the LSTMs with the hypothesis that the availability of more information and the efficient selection of the right information will in-turn result in a better model. The use of multiple memory cells is concealed within the nodes so as to keep the network dynamics same as that of standard LSTM networks. This introduces the need to have an efficient selection mechanism that selects a single value from the multi-cells such that a single output is transmitted from the node.\\

Concretely, our contributions are two-fold:
\begin{itemize}
	\item We propose a multi-cell node architecture for LSTM network and investigate the optimum strategies for selecting a particular cell value or combining all the memory cell values so as to output a single value from the node.
    \item Further, we apply the multi-cell LSTM for neural language modeling and compare its performance with the state-of-the-art Zaremba's models \cite{DBLP:journals/corr/ZarembaSV14}.
\end{itemize}

\section{Neural Language Modeling}

Neural Language Modeling came into limelight with the  `Neural Probabilistic Language Model' (hereafter referred as NPLM) proposed by Bengio et al. \cite{Bengio:2003:NPL:944919.944966}. It deals with the challenges of n-gram language models and the `Curse of dimensionality' \cite{curseof_Bengio:2000:TCD:2325773.2326573} by simultaneously learning the distributed representation of words and the probability distribution for the sequences of words expressed in terms of these representations. The feed-forward neural network based NPLM architecture makes use of a single hidden layer with hyperbolic tangent activation to calculate the probability distribution. The softmax output layer gives the probabilities. Generalization is obtained with the use of distributed word representations. Arisoy et al. \cite{Arisoy:2012:DNN:2390940.2390943} extended Bengio's model by adding more hidden layers (upto 4). They observe that the deeper architectures have the potential to improve over the  models with single hidden layer.  \\

NPLM was exceptionally successful and the further investigations \cite{ClassLM_DBLP:journals/corr/cs-CL-0108005} show that this single model outperforms the mixture of several other models which are based on other techniques. Even though Bengio et al. were successful in incorporating the idea of word similarities into the model \cite{Bengio:2003:NPL:944919.944966}; owing to the use of fixed length context that needs to be specified prior to the training; NPLM was unable to make use of larger contexts. In contrast to the feed forward neural networks, the recurrent neural networks do not limit the context size. With recurrent connections, theoretically,  information can cycle in the network for arbitrarily long time. In \cite{Mikolov} Mikolov et al. study the `Recurrent Neural Network based Language Model' (hereafter referred as RNNLM). RNNLM makes use of `simple recurrent neural network' or the `Elman network' architecture \cite{Elman_COGS:COGS203}. The network consists of an input layer, a hidden layer (also known as the context layer) with sigmoid activation and the output softmax layer. Input to the network is the concatenation of one-hot-vector encoding of the current word and the output from neurons in context layer at previous time-step. Learning rate annealing and early stopping techniques are used for efficient training. \\  

Even though the preliminary results on toy tasks were promising, Mikolov et al. \cite{Mikolov} conclude that it is impossible for simple RNNs trained with gradient descent to truly capture long context information, mainly because of the `Vanishing gradient' and `Exploding gradient' problems. One way to deal with this inefficacy of gradient descent approach to learn long-range context information in the simple RNN is to use an enhanced learning technique. Martens et al. \cite{Martens2011LearningRN} have shown that the use of HF optimization technique can solve the vanishing gradient problem in RNNs. In \cite{ICML2011Sutskever_524} Sutskever et al. propose a character-level language model, making use of the Hessian-free optimization to overcome the difficulties associated with RNN training.\\

Another popular technique to solve the training difficulties with the simple RNN is to modify the network to include the `memory units' that are specially structured to store information over longer periods. Such networks are termed as Long Short Term Memory (LSTM) networks. In \cite{DBLP:journals/corr/ZarembaSV14} Zaremba et al. makes use of a LSTM network for language modeling (hereafter referred as LSTMLM). LSTMLM uses two LSTM layers as the hidden network. As in the case of \cite{Bengio:2003:NPL:944919.944966}, the  network simultaneously learns the distributed word embeddings and the probability function of the sequences of words represented in terms of these embeddings. Softmax output layer gives the probability distribution. The network is trained by means of back-propagation and stochastic gradient descent. Learning rate annealing is used for the efficient training. `Gradient Clipping' \cite{gradClip_Pascanu:2013:DTR:3042817.3043083} is employed to prevent the `exploding gradient' problem. Zaremba et al. were also successful in identifying and employing an efficient way to apply dropout regularization to the RNNs. Because of this, the dropout-regularized LSTMLM was able to perform much better than the non-regularized RNN based language models.\\

Our model is similar to the neural language model by Zaremba et al. \cite{DBLP:journals/corr/ZarembaSV14}. Instead of the standard single cell LSTMs used by Zaremba et al., we use our multi-cell LSTMs.

\section{Node structure in LSTM and the proposed multi-cell LSTM node}

\begin{figure}[!ht]
  
  \centering
  \includegraphics[width=\linewidth]{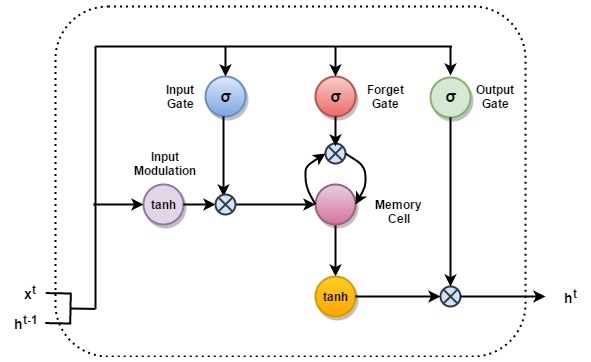}
\caption{LSTM cell structural diagram }
\label{lstm_node}  
\end{figure}

Figure 1 shows the structure of a LSTM node. Core of the LSTM node is the memory cell which stores its state. Information stored in the memory cell is updated with the help of two gates. The `Input gate' regulates the amount of information that gets added into the cell state, whereas the `Forget gate' controls the amount of information that is removed from the current cell state. The input, forget and output gates use the sigmoid activation. The hyperbolic tangent  of the cell state, controlled by the `Output gate' is given as the output of the node.

\begin{equation}
	{a^t} = \tanh{(W_cx^t + U_ch^{t-1} + b_c)}
\end{equation}
\begin{equation}
	{i^t} = \sigma{(W_ix^t + U_ih^{t-1} + b_i)}
\end{equation}
\begin{equation}
    {f^t} = \sigma{(W_fx^t + U_fh^{t-1} + b_f)}
\end{equation}
\begin{equation}
    {o^t} = \sigma{(W_ox^t + U_oh^{t-1} + b_o)}
\end{equation}
\begin{equation}
    {c^t} = i^t.a^t + f^t.c^{t-1}
\end{equation}
\begin{equation}
	{h^t} = o^t . \tanh{(c^t)}
\end{equation}

The equations (1) to (6) describe how a LSTM cell is updated at every time-step $t$. In these equations, $x^t$ is the input to the memory cell at time $t$. $i^t$, $f^t$ and $o^t$ are the input, forget and output gate values at time $t$. $a^t$ is the modulated input to the cell at time $t$. $c^t$ denotes the cell state value at time $t$, where as $c^{t-1}$ is the cell state value in the previous time step. $h^t$ is the output of the cell at time $t$.\\

Because of this particular node structure, LSTMs are able to store the information for longer time-steps and hence can solve the vanishing gradient problem. This makes LSTMs a popular and powerful choice for langauge modeling applications, where there is a need to store long contextual information. However, the `exploding gradients' problem still prevails in the network. 'Gradient clipping' \cite{gradClip_Pascanu:2013:DTR:3042817.3043083} can be used to prevent this.\\

Figure 2 gives the architecture of a multi-cell LSTM node. Instead of a single memory cell in a standard LSTM node, each multi-cell LSTM node holds $m$ memory cells. The gate values and the input to the cells are calculated as in the case of standard LSTM node. These are then broadcast to the memory cells and each cell updates its values using equation (7). 

\begin{equation}
    {c^t_i} = i^t.a^t + f^t.c_i^{t-1}
\end{equation}
Where $c_i^t$ denotes the content of memory-cell $i$ at time $t$. $c_i^{t-1}$ is the content of the memory-cell at time $t-1$.\\

Updated cell values flow into the selection module, where they are combined or a single value is selected based on the underlying strategy. Regulated by the output gate, the hyperbolic tangent of the selected/combined cell value moves into output. The number of memory cells in each node, $m$ is a hyper-parameter of the network that can be tuned while training. In the next section, we discuss various strategies that can be used for the selection of a single value from the multi-cells.\\

\begin{figure}[!ht]
  
  \centering
  \includegraphics[width=\linewidth]{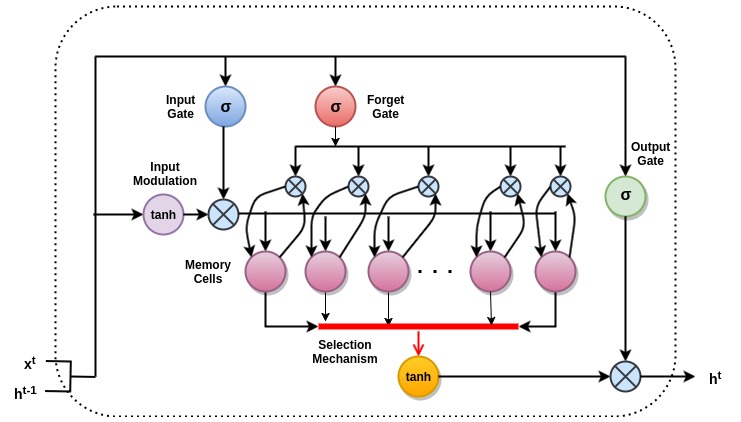}
\caption{Multi-cell LSTM node }
\label{chap3:multi-cell_node}  
\end{figure}

\section{Strategies for combining multiple memory cell values in a multi-cell LSTM node}

As in the case of standard LSTM node, mutli-cell LSTM node takes two inputs ($x^t$, $h^{t-1}$) and provides a single output ($h^t$). To get the single output from multiple memory cells, we need to have a selection mechanism. Here we discuss the different strategies applied to select a particular memory cell/combine all the cell values, so as to pass a single value to the output. From the selection module, we obtain the effective cell value ($c_{eff}$). Output of the node is then computed using Equation 8.

\begin{equation}
	{h^t} = o^t . \tanh{(c_{eff}^t)}
\end{equation}

\subsection{Simple Mean}

The effective cell value, $c_{eff}$ is calculated as the mean value of the multiple memory cells in the node. Hyperbolic tangent of $c_{eff}$ regulated by the output gate forms the output of the node.
\begin{equation}
	{c_{eff}^t} = \frac{1}{m}\sum_{i=1}^{m} c_i^t 
\end{equation}
where $m$ is the number of memory cells in the node. $c_i^t$ denotes the content of memory-cell $i$ at time $t$. 

\subsection{Weighted Sum}

Each cell is given a static weight. First cell is given the maximum weighting and the other cells are weighted in decreasing order with a constant decay from the previous cell weight. $c_{eff}$ is calculated as the sum of these weighted cells.
\begin{equation}
	{c_{eff}^t} = \sum_{i=1}^{m} c_i^t.w_i 
\end{equation}
where $w_i$ is the static weight for the $i^{th}$ memory cell. 

\subsection{Random Selection}

From the $m$ memory cells available, we randomly pick a memory cell. This is our effective cell value ($c_{eff}$). 
\begin{equation}
	{c_{eff}^t} = Random(C^t) 
\end{equation}
where $C^t$ is a one dimensional vector holding the values of all memory cells in the node at time $t$.

\subsection{Max Pooling}

Similar to the `max pooling' layer used in Convolutional Neural Networks (CNN), selection mechanism of the node picks the maximum value stored in its memory cells. This is the effective cell value of the node. 

\begin{equation}
	{c_{eff}^t} = Max(C^t) 
\end{equation}

\subsection{Min-Max Pooling}

We define a threshold value for output gate.  If the value is less than the threshold, we take the minimum of the cell values. We go with the maximum of cell values, if the gate value is greater than the threshold. 

\begin{equation}
	{c_{eff}^t} = 
    \begin{cases}
    		Min(C^t), & \text{if}\ o^t < o_{thr}\\
			Max(C^t), &  \text{otherwise}
	\end{cases} 
\end{equation}

where $o_{thr}$ is the threshold value for the output gate. It is a hyper parameter that can be tuned by training.
  
\subsection{Learnable Weights for the cells }

We associate each memory cell with a trainable weight that gets updated in the backward pass. Now we can combine all the weighted cells or select a particular cell by applying any of the above strategies. Here we demonstrate it with the `Max Pooling' technique. Maximum of the weighted cell values is taken as the effective cell value.

\begin{equation}
	{c_{eff}^t} = Max(C^t.W_{cell}^t) 
\end{equation}
where $W^t_{cell}$ is the learned weights for the memory cells at time $t$.

\section{Network Architecture}

Architecture of our network is same as the one used by Zaremba et al. \cite{DBLP:journals/corr/ZarembaSV14} for language modeling. Input is given through an embedding layer having the same size as that of the hidden layers. It is followed by two hidden layers comprising of LSTM nodes. Fully connected softmax layer gives the output probability distribution.

\section{Results and Observations}

Here we discuss in detail about the datasets, experimental setup, results obtained and the comparisons with the results of Zaremba et al.  \cite{DBLP:journals/corr/ZarembaSV14}.

\subsection{Dataset}

Penn Tree Bank, popularly known as the PTB dataset \cite{Marcus:1994:PTA:1075812.1075835} was used for the experiments. Originally a corpus of English sentences with linguistic structure annotations, it is a collection of 2,499 stories sampled from the three year Wall Street Journal (WSJ) collection of 98,732 stories. A variant of the original dataset, distributed with the Chainer package \cite{chainer_learningsys2015} for Python was used for our experiments. The corpus with a vocabulary of 10000 words is divided into training, test and validation sets consisting of 929k, 82k and 73k words respectively.

\subsection{Evaluation Metric}

Perplexity measure was used as the performance evaluation metric. Most popular performance metric for the language modeling systems, Perplexity is a measure of how well the probability model or the probability distribution predicts a sample. Lower the Perplexity, better the model is. A low perplexity indicates that the model is good at predicting the sample. The perplexity of a probability distribution is defined as follows:
\begin{equation}
Perplexity = 2^{H(p)} = 2^{-\sum_{x}p(x)log_2p(x) }
\end{equation}
where, $H(p)$ is the entropy of the distribution and the $x$ ranges over the events.
\subsection{Experimental Setup}

Multi-cell LSTM based language models (hereafter referred as MLSTM-LM) of three different sizes were trained. These are denoted as the small MLSTM-LM, medium MLSTM-LM and the large MLSTM-LM. All of these have two layers of multi-cell LSTMs and are unrolled for 35 steps. We follow the conventions used by Zaremba et al. in \cite{DBLP:journals/corr/ZarembaSV14}. Hidden states are initialized to 0. We also ensure the statefulness of the model by which the final hidden state of the current mini-batch is used as the initial hidden state of the subsequent mini-batch. Mini-batch size is kept as 20. As in \cite{DBLP:journals/corr/ZarembaSV14}, we apply dropout on all the connections other than the recurrent connections in LSTM layers. \\ 

The small MLSTM-LM has 200 nodes per hidden layer. Dropout rate of 0.4 is used for these networks. Initial learning rate of 1 is used. The medium MLSTM-LM has 650 nodes in the hidden layers. Dropout rate of 0.5 and initial learning rate of 1.2 were used. For the large MLSTM-LMs having 1500 nodes per hidden layer, initial learning rate of 1.2 and dropout rate of 0.65 were used. All the models use the Algorithm \ref{lrdecay} for learning rate annealing.\\

At the end of each epoch, perplexity is computed for the validation set. This is used to anneal the learning rate. Learning rate annealing algorithm is run after each epoch. We define \textit{`epochs to wait'} as the number of epochs to wait for an improvement in the perplexity, before annealing the learning rate. \textit{`minimum reduction'} is the threshold improvement in perplexity from the previously recorded perplexity value to consider an improvement. \textit{`minimum learning rate'} defines the lower bound for learning rate annealing. \textit{`given chances'} holds the number of epochs since last improvement in the metric. If there is no considerable improvement for perplexity from the previous epoch, \textit{`given chances'} is incremented. If \textit{`given chances'} crosses the \textit{`epochs to wait'}, we reduce the learning rate by multiplying it with the decay factor.

    \begin{algorithm}[!ht]    
		\caption{Procedure: Learning rate annealing}\label{lrdecay}          
        \DontPrintSemicolon
        \KwIn{Learning rate decay, minimum learning rate, minimum reduction, epochs to wait}
        \KwOut{updated learning rate}
		\Begin{
        	\eIf{current perplexity is greater than (previous perplexity- minimum reduction)}
            {\eIf{given-chances less than the epochs to wait}
            	{Increment given-chances.}
               {new learning rate = max(minimum learning rate, learning rate * learning rate decay)\\ given-chances =0\\ }}
            {given-chances = 0} 
		Return updated learning rate	
		}        
    \end{algorithm}
We define \textit{'Learning rate decay'} as 0.5, \textit{'epochs to wait'} as 2, \textit{'minimum reduction'} as 2 and \textit{'minimum learning rate'} as 0.0001.

\subsection{Results and Discussion}

Experiments were conducted with all three models, choosing different selection strategies listed in Section 4. Hyper parameters; including the number of memory cells per node; were fine-tuned with the help of repeated runs. Table \ref{chap3:comaparisonSelectionStrategies} lists the performance of Large MLSTM-LM under different selection strategies. All these models have 10 memory cells per LSTM node. As we can see, the selection strategies namely Simple Mean, Random Selection, Max Pooling and the Min-Max Pooling gives almost same results, whereas the other two strategies; Weighted Sum and Learnable Weights for the cells; do not give comparable results. We obtain the top performing model when the `Max Pooling' strategy is used.

\newcolumntype{L}[1]{>{\hsize=#1\hsize\raggedright\arraybackslash}X}%
\newcolumntype{R}[1]{>{\hsize=#1\hsize\raggedleft\arraybackslash}X}%
\newcolumntype{C}[2]{>{\hsize=#1\hsize\columncolor{#2}\centering\arraybackslash}X}%

\begin{table}[!hb]
\centering

\begin{tabularx}{\columnwidth}{  L{0.6}  R{0.3}  R{0.3} }
  \toprule
  Selection Strategy & Validation Perplexity & Test Perplexity \\
  \midrule
Simple Mean & 81.27 & 77.6 \\
Random Selection & 81.52 & 77.46 \\
Max Pooling & 80.62 & 77.12 \\
Min-Max Pooling & 81.12 & 77.53 \\
  \bottomrule
\end{tabularx}
\caption{Comparison of the multi-cell selection strategies}
\label{chap3:comaparisonSelectionStrategies}
\end{table}

\begin{table}[!hb]
\centering

\begin{tabularx}{\columnwidth}{  L{0.6}  R{0.3}  R{0.3} }
  \toprule
  Number of Memory-Cells & Validation Perplexity & Test Perplexity \\
  \midrule
2 & 81.41 & 77.7 \\
3 & 81.18 & 77.32 \\
4 & 80.89 & 77.47 \\
5 & 81.42 & 77.9 \\
6 & 81.17 & 77.17 \\
7 & 80.99 & 77.43 \\ 
8 & 81.16 & 77.43 \\
9 & 81.72 & 77.62 \\
10 & 80.62 & 77.12 \\
15 & 81.71 & 77.56 \\
30 & 81.61 & 77.3 \\
50 & 81.49 & 77.69 \\
  \bottomrule
\end{tabularx}
\caption{Model performance on varying the number of memory cells in the nodes}
\label{chap3:comaparisonMemoryCellSize}
\end{table}

Table \ref{chap3:comaparisonMemoryCellSize} presents the experiments to choose the optimum number of memory cells. All these experiments were carried out with the Large MLSTM-LM network, using `Max Pooling' strategy. Models perform more or less in a similar fashion despite the change in number of memory cells. We recorded the lowest perplexity with the model having 10 memory cells per LSTM node.\\

\begin{table}[!hb]
\centering

\begin{tabularx}{\columnwidth}{  L{1.5}  R{0.6}  R{0.4} R{0.4} }
  \toprule
  Model & Number of nodes in LSTM layers & Validation Perplexity & Test Perplexity \\
  \midrule
  \multicolumn{4}{l}{Results by Zaremba et al. \cite{DBLP:journals/corr/ZarembaSV14}}
  \\
  \midrule
  Non-regularized LSTM & 200 &	120.7 & 114.5\\
Medium-regularized LSTM & 650 &	86.2 & 82.7\\
Large regularized LSTM & 1500 &	82.2 & 78.4\\
\midrule
\multicolumn{4}{l}{Top models from experiments} \\
\midrule
Small MLSTM-LM & 200 & 94.5 & 89.39\\
Medium MLSTM-LM & 650 & 83.88 & 79.95\\
Large MLSTM-LM & 1500 & 80.62 & 77.12\\
Large regularized LSTM(Replicated) & 1500 & 81.87 & 77.85\\
Regularized LSTM(Replicated) & 200 & 103.216 & 99.19\\
  \bottomrule
\end{tabularx}
\caption{Comparison of the top models with the results of Zaremba et.al.
}
\label{chap3:comaparisonTopModels}
\end{table}

Table \ref{chap3:comaparisonTopModels} presents the performance of top models. As evident from the table, MLSTM-LM models outperform the models from \cite{DBLP:journals/corr/ZarembaSV14} on both validation and test sets. While replicating the test of Zaremba et al. using our learning rate annealing algorithm; instead of their original strategy; we obtain better results as compared to the results reported by them in \cite{DBLP:journals/corr/ZarembaSV14}.\\ 

\begin{figure}[!ht]
\centering     
 \pgfplotsset{width=20cm,compat=1.9}
\begin{tikzpicture}[scale=0.55]
	\begin{axis}[
		title={Epoch v/s Validation Perplexity Plot},
		height=10cm,
		width=12.5cm,
		ymajorgrids=true,
   		 grid style=dashed,
		xlabel = Epochs,
		ylabel = Validation Perplexity
	]
	\addplot table[x index = 0,y index=1,col sep=comma] {dataGraph/orig_best/textgen_363.dat};
	\addlegendentry{MLSTM-LM}
	
	\addplot table[x index = 0,y index=1,col sep=comma] {dataGraph/orig_best/textgen_404.dat};
	\addlegendentry{LSTM-LM}
	\end{axis}
\end{tikzpicture}
     

\caption{Epoch v/s Validation Perplexity plot for LSTM-LM and MLSTM-LM}
\label{chap3:mlstm-lstm}
\end{figure}
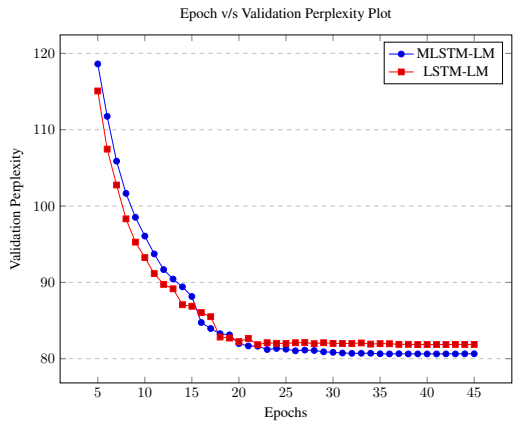

\begin{figure}[!ht]
\centering     

 \pgfplotsset{width=10cm,compat=1.9}
\begin{tikzpicture}[scale=0.55]
	\begin{axis}[
		title={Epoch v/s Validation Perplexity Plot},
		height=10cm,
		width=12.5cm,
		ymajorgrids=true,
   		 grid style=dashed,
		xlabel = Epochs,
		ylabel = Validation Perplexity
	]

	\addplot table[x index = 0,y index=1,col sep=comma] {dataGraph/threeModl/textgen_363.dat};
	\addlegendentry{Large MLSTM-LM}
	
	\addplot table[x index = 0,y index=1,col sep=comma] {dataGraph/threeModl/textgen_379.dat};
	\addlegendentry{Medium MLSTM-LM}
	
    \addplot table[x index = 0,y index=1,col sep=comma] {dataGraph/threeModl/textgen_67.dat};
	\addlegendentry{Small MLSTM-LM}
    
	\end{axis}
\end{tikzpicture}

\caption{Epoch v/s Validation Perplexity plot for Small, Medium and Large MLSTM-LMs}
\label{chap3:graph3models}
\end{figure}
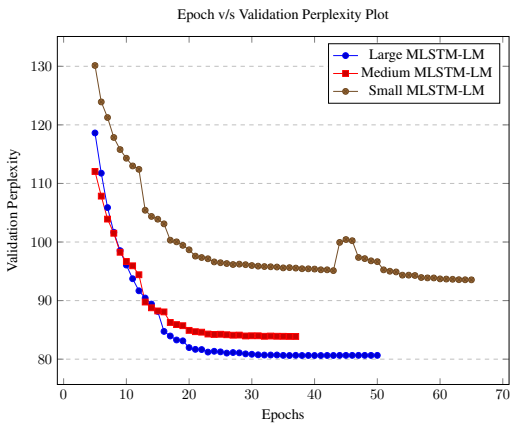

Figure \ref{chap3:mlstm-lstm} gives the Epoch v/s Validation set perplexity plot for the top MLSTM-LM and replicated result of Large regularized LSTM model by Zaremba et al. \cite{DBLP:journals/corr/ZarembaSV14}. Figure \ref{chap3:graph3models} is the Epoch v/s Validation set perplexity plot for small, medium and large MLSTM-LM models. While Zaremba et al. \cite{DBLP:journals/corr/ZarembaSV14} reports the saturation of large model at Epoch 55 and the medium model at Epoch 39, all of our models attain saturation well before that. All of our models were able to outperform the models from \cite{DBLP:journals/corr/ZarembaSV14} with a training time of approximately 30 epochs. Interestingly the replication of Zaremba's experiment on the large regularized LSTM model using our learning rate annealing algorithm also saturates early. This also shows the effectiveness of our annealing algorithm.  

\subsection{Miscellaneous Experiments}

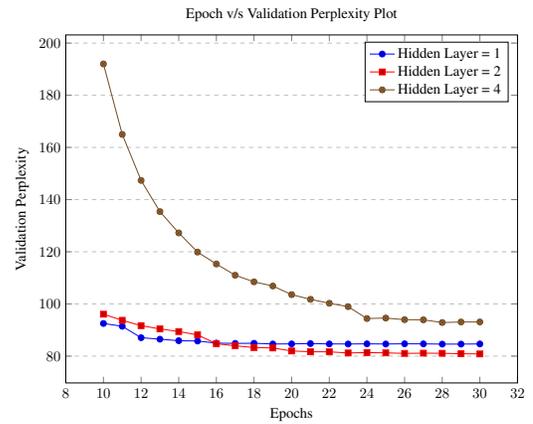
\begin{figure}[!ht]
\centering     

 \pgfplotsset{width=10cm,compat=1.9}
\begin{tikzpicture}[scale=0.55]
	\begin{axis}[
		title={Epoch v/s Validation Perplexity Plot},
		height=10cm,
		width=12.5cm,
		ymajorgrids=true,
   		 grid style=dashed,
		xlabel = Epochs,
		ylabel = Validation Perplexity
	]

	\addplot table[x index = 0,y index=1,col sep=comma] {dataGraph/hidden_layer/textgen_314.dat};
	\addlegendentry{Hidden Layer = 1}
	
	\addplot table[x index = 0,y index=1,col sep=comma] {dataGraph/hidden_layer/textgen_363.dat};
	\addlegendentry{Hidden Layer = 2}
	
    \addplot table[x index = 0,y index=1,col sep=comma] {dataGraph/hidden_layer/textgen_319.dat};
	\addlegendentry{Hidden Layer = 4}
    
	\end{axis}
\end{tikzpicture}
\caption{Epoch v/s Validation Perplexity plot for MLSTM-LMs with different number of hidden layers}
\label{chap3:hiddenLayerComparison}
\end{figure}

Experiments were also conducted to check the performance of the models on varying number of hidden layers and embedding size. Table \ref{chap3:comaparisonHiddenLayersTable} summarizes the experiments with different number of hidden layers. Figure \ref{chap3:hiddenLayerComparison} gives the Epoch v/s Validation perplexity plot for the models. Even though we have a fairly good result with a single hidden layer model, we obtain the optimum result with the network having two hidden layers. Performance deteriorates when more hidden layers are added. It might be an indication of the requirement for better regularization techniques as we add more hidden layers. More research is required on this and we leave it for future work.

\begin{table}[!hb]
\centering
\begin{tabularx}{\columnwidth}{  L{0.6}  R{0.3}  R{0.3} }
  \toprule
  Hidden Layers & Validation Perplexity & Test Perplexity \\
  \midrule
1 & 84.50 & 80.91 \\
2 & 80.62 & 77.12 \\
4 & 91 & 87.2 \\
  \bottomrule
\end{tabularx}
\caption{Model performance on varying the number of hidden layers}
\label{chap3:comaparisonHiddenLayersTable}
\end{table}

\begin{table}[!hb]
\centering
\begin{tabularx}{\columnwidth}{  L{0.4}  R{0.3}  R{0.5} R{0.6} R{0.5} }

\toprule
Model & Units & Embedding size & Validation Perplexity & Test Perplexity\\
\midrule
Small & 200 & 300 & 127.21 & 123\\
Small & 200 & 600 & 124.51 & 121.4\\
Small & 200 & 200 & 94.5 & 89.39\\
Large & 1500 & 300 & 102.92 & 99\\
Large & 1500 & 600 & 95.07 & 92.38\\
Large & 1500 & 1500 & 80.62 & 77.12\\
  \bottomrule
\end{tabularx}
\caption{Perplexity of the models on varying the embedding size}
\label{chap3:comaparisonEmbeddSize}
\end{table}

Table \ref{chap3:comaparisonEmbeddSize} describes the experiments conducted to study the effect of embedding size on the model performance. Small and Large MLSTM-LMs were used for the study. As evident from the table, both the models give best results when the embedding size equals the number of nodes in the hidden layers. Performance of the models deteriorates as we use embedding size different from the hidden layer size.\\ 

Experiments were also conducted to study the effects of weight-decay and max-norm regularization techniques. Srivastava et al. \cite{Srivastava:2014:DSW:2627435.2670313} suggest that the max-norm regularization which thresholds the L2 norm of the weight vectors between 3 and 4 enhances the performance of dropout-regularized networks. But we were unable to obtain any considerable improvement in the model performance with the max-norm regularization. As reported by Mikolov et al. \cite{Mikolov}, regularization of network to penalize large weights by weight-decay did not provide any significant improvements.

\section{Conclusion}
In this paper, we have proposed the multi-cell LSTM architecture and discussed the various selection strategies to select a particular memory cell or combine all the memory cells of a node. We have applied the model successfully for language modeling. Effectiveness of different selection strategies and the effect of varying the number of memory cells on model performance were also studied. Our MLSTM-LM models were able to outperform the state-of-the-art Zaremba's models and attain saturation early as compared to the results reported by Zaremba et al. \cite{DBLP:journals/corr/ZarembaSV14}.\\

Multi-cell LSTMs need to be investigated more for its performance on the other applications of standard LSTMs. We have explored only the `maximum' and `average' functions for selecting a particular value out of the multi-cells. Further ideas from signal theory can also be used with the use of linear filters, as average is also a linear filter. MLSTM-LMs should be applied on the various language modeling applications like speech recognition in order to measure its practicality.

\bibliographystyle{named}
\bibliography{ijcai18}

\end{document}